\documentclass[letterpaper]{article} %
\usepackage{aaai2027}  %

\usepackage[hyphens]{url}  %
\usepackage{graphicx} %
\usepackage{natbib}  %
\usepackage{caption} %
\usepackage{algorithm}
\usepackage{algorithmic}

\usepackage{newfloat}
\usepackage{listings}
\DeclareCaptionStyle{ruled}{labelfont=normalfont,labelsep=colon,strut=off} %
\floatstyle{ruled}
\newfloat{listing}{tb}{lst}{}
\floatname{listing}{Listing}

\usepackage{booktabs}

\usepackage{tabularx}
\usepackage{amsmath}
\usepackage{amsfonts}
\usepackage{makecell}

\nocopyright %

\newcommand{\method}[0]{{Z-PEFT}}

\title{\method{}: Zero-shot Backdoor Detection in Parameter-Efficient Fine-Tuning via Canonical Spectral Signatures}

\author{
    Nicola Pitzalis \quad Donald Shenaj \quad Giacomo Cignoni  \quad
    Andrea Cossu \\ Davide Bacciu \quad Antonio Carta 
}
\affiliations{
    University of Pisa \\
    pitzalis93@gmail.com, donald.shenaj@di.unipi.it, giacomo.cignoni@phd.unipi.it, \\
    andrea.cossu@unipi.it, davide.bacciu@unipi.it, antonio.carta@unipi.it

}

\begin{document}

\maketitle

\begin{abstract}
Parameter-Efficient Fine-tuned (PEFT) models are frequently downloaded from open repositories by practitioners. This widespread practice creates a significant attack surface, as malicious actors can publish backdoored models that induce specific behaviors in response to predefined triggers. We study the problem of weight-space backdoor detection, where a detector classifier predicts whether a model is malicious using only its weights, enabling a lightweight safety mechanism. Most existing methods are designed and evaluated in a closed-world setting, where the detector is trained and tested on the same attack type. In contrast, we evaluate backdoor detection under novel conditions, including previously unseen attacks and datasets. We propose \method{}, a lightweight meta-classifier that relies exclusively on layer-wise spectral measures for classification. Our experiments show that strong performance in the closed-world setting does not necessarily translate to high accuracy in zero-shot backdoor detection. Among weight-space detectors, \method{} achieves the best performance while maintaining low and scalable computational cost\footnote{Source code will be released upon acceptance.}.
\end{abstract}

\section{Introduction}
Parameter-efficient fine-tuning (PEFT) has emerged as the dominant paradigm for adapting large pre-trained language models to downstream tasks, offering an efficient alternative to full fine-tuning \cite{ding2022delta,han2024parameter}. Rather than updating all parameters, PEFT methods introduce a small trainable components while keeping the rest frozen. 
Among these, low-rank adaptation constitutes a particularly effective family of methods that parameterizes weight updates through compact low-rank decompositions~\citep{hu2021lora,zhang2023adalora,dettmers2023qlora,liu2024dora}.
These methods have enabled cost-effective adaptation of large-scale open-source models, including Llama-2~\cite{touvron2023llama}, Qwen1.5-7B \cite{qwen1.5}, Flan-T5-XL \cite{chung2024scaling}, and RoBERTa \cite{liu2019roberta}, across a wide range of language understanding and generation benchmarks.

\begin{figure}[t]
    \centering
    \includegraphics[width=\linewidth]{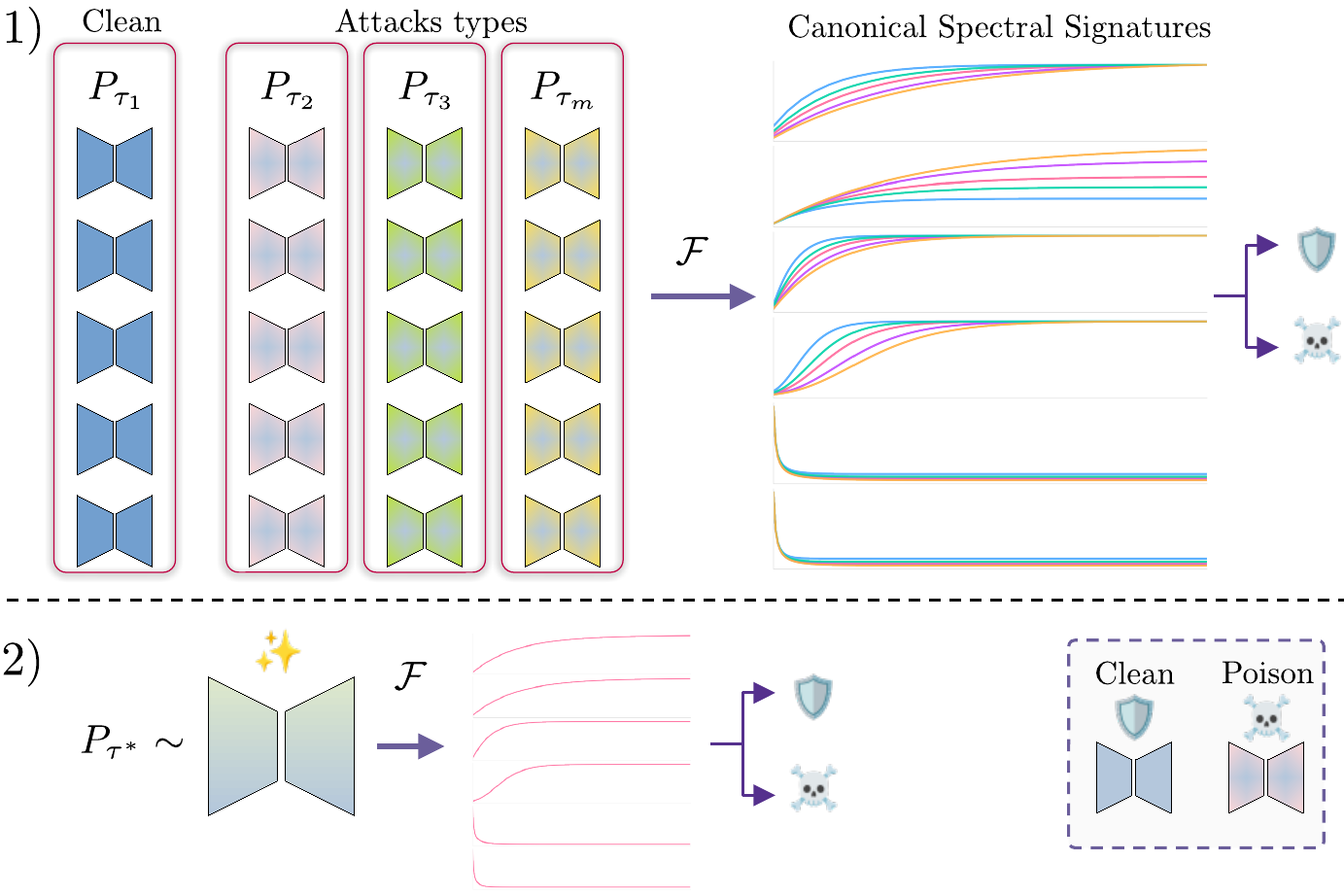}
    \caption{1) Z-PEFT is trained on a set of clean and backdoored PEFT adapters. The method extracts spectral features from weight matrices to compute a cheap representation of the model. 2) At test time, Z-PEFT can detect backdoored models even when they are trained using unknown attacks.}
    \label{fig:teaser}
\end{figure}
As models grow larger and full finetuning becomes impractical, practitioners increasingly rely on community-contributed PEFT adapters hosted on platforms such as Hugging Face rather than training their own.
While this democratizes access to capable models, it also introduces a critical and underexplored security risk: a malicious actor can publish a backdoored adapter that behaves normally on clean inputs but produces targeted, adversarially chosen outputs whenever a specific trigger pattern is present. Because the adapter represents only a small fraction of the total model parameters, such backdoors can be difficult to detect through standard evaluation and may persist even after downstream fine-tuning \cite{li-etal-2024-cleangen}.
Backdoor attacks on language models have been studied across several threat surfaces. Prompt-based attacks \cite{zhao-etal-2023-prompt} embed triggers directly in the input, while more recent work has demonstrated that backdoors can be injected through adapter weights \cite{liu-etal-2025-loratk} or attention mechanisms \cite{lyu-etal-2023-attention}, making them harder to detect without access to model internals. These findings motivate the need for detection methods that operate directly on the adapter parameters rather than on model behavior, which may be indistinguishable from benign under standard evaluation conditions.

In this work, we argue that backdoor detection is intrinsecally an \emph{open-world problem} since the detector at test-time must generalize to novel attacks and datasets. We call this scenario the \emph{zero-shot backdoor detection} problem and we study it via weight-space detection, where the detector has access only to adapter weights, with no knowledge of triggers, training data, or base model behavior.
We propose \method{} to address it (Figure \ref{fig:teaser}). 

Our main contributions are:
\begin{itemize}
\item We propose \method{}, combining head-wise spectral features with a lightweight meta-classifier which can scale to large PEFT repositories.
\item We show that preserving head-wise, projection-wise structure captures backdoor signatures that global spectral summaries miss, at a fraction of the cost of raw tensor-based approaches.
\item We conduct extensive experiments across multiple model families and attack configurations, demonstrating that \method{} outperforms existing detection baselines in the zero-shot setting.
\end{itemize}

\section{Related Work}
Recent work on backdoor detection for PEFT adapters has explored static methods that operate directly on adapter weights, avoiding the need for trigger knowledge or model execution. We discuss two closely related approaches, PEFTGuard~\cite{sun2025peftguard} and weight-space detection for LoRA adapters (WSD)~\cite{puertolas2026weightspace}.

PEFTGuard formulates adapter backdoor detection as a supervised binary classification problem over PEFT parameters. It constructs a tensor from the self-attention adapter weights by combining the query and value updates across layers, and feeds this representation to a convolutional meta-classifier followed by fully connected layers. This design preserves raw structural information from adapter weights, but, because the classifier operates directly on the original high-dimensional weight tensor, its memory and computation usage grow with the number of layers and the size of the selected projections.

WSD follows a more compact spectral approach. It selects a fixed layer, reconstructs the LoRA weight updates, aggregates the attention-projection updates, and extracts a small set of spectral and distributional statistics Features include leading singular values, matrix norms, energy concentration, spectral entropy, and kurtosis, normalized against a benign reference bank and classified using logistic regression. This approach uses a limited set of features from a single layer, reducing training and inference costs after preprocessing at the expense of a more compressed representation.

Our method occupies a middle ground between raw weights and limited spectral descriptors.
It converts adapter weights into fixed-dimensional features while preserving the model hierarchy by extracting spectral statistics separately for each attention head, layer, and projection matrix.
This yields a representation that is more compact and scalable than PEFTGuard’s raw weights, yet structurally richer than limited spectral summaries.
We provide a more detailed comparison in the next Sections.

\section{Zero-shot Backdoor Detection in PEFT}
\label{sec:zero_day_backdoor_detection}

In practical PEFT deployment, the defender may receive an adapter from an external source and must decide whether it is safe to load it into a base model. This differs from standard backdoor detection benchmarks, where the attack family, trigger construction, and data distribution may be known in advance. In realistic settings, the defender may not know which attack was used, which task it targets, or whether any labeled backdoored adapters are available for training. We refer to this broader problem as zero-shot backdoor detection in PEFT: detecting malicious adapters whose attack mechanism may not have been observed during detector training.

This setting also changes the role of the detector. If the deployment-time setting is known in advance, one may train or select a specialized classifier for that specific condition --- for example a classifier specialized to a particular architecture, PEFT method, adapter rank, downstream distribution, or attack family. However, this assumption becomes unrealistic in the zero-shot setting. Since the defender does not know beforehand what kind of adapter will be inspected, there is no reliable way to choose the best specialized detector at test time. This motivates a unified detector that can operate across heterogeneous settings, including generalizing to novel benchmarks and attacks.

Formally, each deployment condition is a separate task \(\mathcal{T}_\tau\) with a data distribution $(\Delta, y) \sim P_\tau$. %
The data distribution is influenced by factors such as the base architecture, the PEFT method, the adapter rank, the downstream data distribution, and the possible attack mechanism. 
Therefore, a realistic detector is one that generalizes to novel tasks, for example by learning task-invariant evidence of backdooring. %

\paragraph{Zero-shot backdoor detection.} In the zero-shot setting, the detector is trained on known tasks but evaluated on at least one unseen task. Formally, training samples come from task distributions \(\{P_{\tau_1}, \ldots, P_{\tau_m}\}\), while evaluation includes a target task \(P_{\tau^\star}\) with
\[
    \tau^\star \notin \{\tau_1,\ldots,\tau_m\}.
\]
The held-out task may differ by attack mechanism, architecture, PEFT method, adapter rank, downstream distribution, or a combination of these factors.

This setting is more realistic because, in practice, the defender usually does not know which attack or adapter configuration will appear in advance. A useful detector should therefore identify backdoor-related weight-space signatures that transfer beyond the specific tasks observed during training. This setting naturally favors multi-task training, since the detector must learn features that are shared across tasks rather than overfitting to one fixed deployment condition.

\section{Methodology}
\label{sec:methodology}

\subsection{Feature representation}
\label{sec:feature_extraction}
Given a PEFT adapter, our goal is to determine whether it is benign or backdoored. The detector operates statically on the adapter weights, requiring neither model execution, input prompts, trigger search, nor access to the original fine-tuning data. We therefore formulate backdoor detection as a binary classification problem in weight space.

Our approach is motivated by the hypothesis that backdoor adapter-tuning introduces localized and structured modifications to the weights~\cite{wu2021adversarial}. Malicious behavior is usually highly specific and is often learned via few high-influence samples. Furthermore, backdoors must override any previous behavior of the model. By contrast, clean tasks usually augment the model with less drastic changes and more natural behavior. As a result, backdoored adapters are expected to cause different weight distributions compared to clean ones. 

We hypothesize that a representation that extracts the spectral features of individual attention heads can discriminate between clean and backdoored models.
Under this hypothesis, it is unnecessary to model the full set of parameter interactions within the adapter. Instead, we represent each attention head by a compact summary of its singular-value spectrum, which captures the dominant structural properties of the corresponding weight update while avoiding the complexity of the full weight matrices. In fact, we argue that using spectral features can help generalize to novel attacks by making it harder for the model to learn attack-specific and trigger-specific patterns in the raw weight space that would not generalize to novel attacks.

For each transformer layer \(l\), we consider the effective LoRA updates to the query and value projections. Given the LoRA factors \(A\) and \(B\), the update added to a frozen weight matrix is
\[
    \Delta W^{(l)} = B^{(l)}A^{(l)}.
\]
Accordingly, for \(p\in\{q,v\}\), we reconstruct
\[
    \Delta W_p^{(l)} = B_p^{(l)}A_p^{(l)}.
\]
For AdaLoRA adapters, the scaling vector \(E\) is absorbed into one of the factors before forming the effective update.

To preserve localized spectral patterns, each projection is partitioned according to its attention heads. Let \(B_{p,h}^{(l)}\) denote the rows of \(B_p^{(l)}\) associated with head \(h\). The corresponding head-wise update is then
\[
    \Delta W_{p,h}^{(l)}
    =
    B_{p,h}^{(l)}A_p^{(l)}.
\]
For each layer \(l\), projection type \(p\), and attention head \(h\), we
extract the singular-value spectrum of the corresponding update block:
\[
    \boldsymbol{\sigma}_{p,h}^{(l)}
    =
    \operatorname{svdvals}\!\left(
        \Delta W_{p,h}^{(l)}
    \right).
\]
We summarize this spectrum using a fixed-dimensional mapping
\(\psi\), obtaining the local spectral descriptor
\[
    \mathbf{s}_{p,h}^{(l)}
    =
    \psi\!\left(\boldsymbol{\sigma}_{p,h}^{(l)}\right).
\]
The representation of adapter \(\mathcal A\) is then formed by concatenating
all local descriptors according to a predetermined ordering of layers,
projection types, and attention heads:
\[
    \mathbf{z}(\mathcal A)
    =
    \operatorname{concat}_{l,p,h}
    \mathbf{s}_{p,h}^{(l)}.
\]
This fixed ordering ensures that corresponding dimensions have the same
interpretation across adapters.

Rather than classifying directly from the complete adapter parameters, we instead learn
\[
    P(y\mid\mathcal A)
    \approx
    P\bigl(y\mid \mathbf{z}(\mathcal A)\bigr),
\]
where \(y\in\{0,1\}\) denotes whether the adapter is benign or backdoored.

The descriptor is defined as
\[
\psi(\boldsymbol{\sigma})=
\Big[ \boldsymbol{\sigma}_{1:k}, E, \|\boldsymbol{\sigma}\|_2, \|\boldsymbol{\sigma}\|_1, r_s, H, r_{\mathrm{eff}}, C, \kappa \Big],
\]
where the components are:

\medskip
\noindent
\textbf{Leading singular values} ($\boldsymbol{\sigma}_{1:k}$): retain the dominant spectral components explicitly. The vector is zero-padded whenever fewer than $k$ singular values are available. Throughout this work, we use $k=8$.

\medskip
\noindent
\textbf{Energy} ($E=\sum_i \sigma_i^2$): measures the overall magnitude of the weight update.

\medskip
\noindent
\textbf{$\ell_2$ norm} ($\|\boldsymbol{\sigma}\|_2=\sqrt{\sum_i \sigma_i^2}$): equivalent to the Frobenius norm of the corresponding weight block, providing a measure of the total update strength.

\medskip
\noindent
\textbf{$\ell_1$ norm} ($\|\boldsymbol{\sigma}\|_1=\sum_i |\sigma_i|$): measures the total spectral mass and complements the energy-based features.

\medskip
\noindent
\textbf{Stable rank} ($r_s=\frac{\sum_i \sigma_i^2}{\sigma_1^2}$): estimates the effective dimensionality of the spectrum. Lower values indicate that the update is concentrated in fewer spectral directions.

\medskip
\noindent
\textbf{Spectral entropy} ($H=-\sum_i p_i\log p_i,\;
p_i=\frac{\sigma_i}{\sum_j \sigma_j}$): quantifies how uniformly the spectral mass is distributed. Lower entropy indicates a more concentrated spectrum.

\medskip
\noindent
\textbf{Effective rank} ($r_{\mathrm{eff}}=\exp(H)$): provides an entropy-based estimate of the number of significant spectral
directions.

\medskip
\noindent
\textbf{Concentration} ($C=\frac{\sigma_1}{\sum_i \sigma_i}$): measures the fraction of spectral mass captured by the leading singular value. High values indicate that the update is dominated by a single direction.

\medskip
\noindent
\textbf{Excess kurtosis} ($\kappa=\frac{\mathbb{E}[(\boldsymbol{\sigma}-\mu)^4]}{\operatorname{Var}(\boldsymbol{\sigma})^2}-3$): measures the peakedness of the singular-value distribution. High kurtosis indicates that the spectral mass is concentrated in a small number of unusually large singular values.

\subsection{Comparison with Prior Static Representations}
\label{subsec:comparison_prior_representations}

The proposed representation differs from prior static detectors mainly in the size and organization of the classifier input. PEFTGuard~\cite{sun2025peftguard} uses the raw weight updates of query and value matrices. For each layer, it concatenates $\Delta W_q^{(l)}, \Delta W_v^{(l)} \in \mathbb{R}^{d \times k}$ into a tensor of size $\mathbb{R}^{2 \times d \times k}$.
Stacking this tensor across \(L\) layers gives an input of size
$\mathbb{R}^{(2L) \times d \times k}$,
so the representation scales as $O(Ldk)$. This preserves the most amount of information, but memory grows with the full size of the selected adapter matrices. It is also unclear whether this is the optimal strategy to generalize across tasks.

Our method instead partitions each projection into \(H\) attention-head blocks and summarizes each block using a fixed-length spectral feature vector containing \(f\) descriptors. Consequently, each head replaces its corresponding \(\frac{dk}{H}\) raw matrix entries with only \(f\) spectral values. For a dataset of \(N\) adapters, the resulting classifier input occupies \(O(NLPHf)\) values, compared with \(O(NLPdk)\) values in PEFTGuard. This corresponds to a reduction in representation size by a factor of $\frac{dk}{Hf}$, since each head is represented by a compact spectral summary rather than its complete weight block. Thus, the classifier operates on a substantially more space-efficient representation while preserving the organization of the adapter by layer, projection type, and attention head.
For example, in a Llama-7B model (\(d=k=4096\), \(H=32\)) with \(f=16\) spectral features per head, each attention head replaces \(\frac{4096 \times 4096}{32} = 524{,}288\) raw matrix entries with only \(16\) spectral features. Across the entire projection, this reduces the representation from \(16{,}777{,}216\) raw values to \(512\) spectral descriptors, corresponding to a compression factor of \(32{,}768\times\).
When accounting for the three projection representations used by our method, compared with the two raw projections retained by PEFTGuard, the overall compression factor is approximately \(21{,}845\times\).

Like WSD, \method{} avoids feeding raw adapter weights directly to the classifier. However, the two representations scale differently. WSD operates exclusively on layer 21 and always produces five summary features; consequently, its classifier input has a fixed dimensionality, independent of the model depth and adapter rank. In contrast, \method{} retains information from multiple layers and projections, producing a richer representation whose dimensionality scales with the model components analyzed while remaining substantially smaller than the raw adapter tensors.

\paragraph{Memory-efficient feature extraction}
Although our features are extracted from \(\Delta W_{p,h}^{(l)}\), Z-PEFT never needs to compute it explicitly. Because the update is low rank, its nonzero singular values can be recovered directly from the LoRA factors. For notational simplicity, consider a generic head-wise factorization
\[
    \Delta W_h = B_hA.
\]
Using the reduced QR decompositions
\[
    B_h = Q_bR_b,
    \qquad
    A^\top = Q_aR_a,
\]
we obtain
\[
    \Delta W_h
    =
    Q_bR_bR_a^\top Q_a^\top
    =
    Q_b\left(R_bR_a^\top\right)Q_a^\top.
\]
Since \(Q_b\) and \(Q_a\) have orthonormal columns, multiplication by these matrices does not change the nonzero singular values. Therefore,
\[
    \operatorname{svdvals}(\Delta W_h)
    =
    \operatorname{svdvals}(R_bR_a^\top),
\]
up to any additional zero singular values. The dimensions of \(R_bR_a^\top\) are governed by the LoRA rank rather than by the input and output dimensions, making the spectral computation substantially more efficient.

We apply this procedure independently to the query and value projections. When their dimensions are compatible, we additionally consider the combined update
\[
    \Delta W_{q+v,h}^{(l)}
    =
    \Delta W_{q,h}^{(l)}
    +
    \Delta W_{v,h}^{(l)}.
\]
This sum retains a low-rank factorization,
\[
    \Delta W_{q+v,h}^{(l)}
    =
    \begin{bmatrix}
        B_{q,h}^{(l)} & B_{v,h}^{(l)}
    \end{bmatrix}
    \begin{bmatrix}
        A_q^{(l)}\\
        A_v^{(l)}
    \end{bmatrix},
\]
so its spectrum can be computed using the same reduced procedure.

\subsection{Backdoor Detection}
The extracted feature vectors are standardized using statistics computed on the training split and used to train a logistic regression classifier with an $\ell_2$ regularization penalty. Logistic regression was selected as a simple linear classifier to evaluate the discriminative power of the proposed representation, avoiding additional modeling capacity that could obscure the contribution of the features themselves. Additionally, its interpretable linear decision function enables a direct analysis of feature coefficients, providing insight into the contribution of each spectral descriptor to the detection decision.

The regularization strength is selected using \(k\)-fold cross-validation on the training set, maximizing the mean area under the receiver operating characteristic curve (AUROC).

Performance is primarily assessed using AUROC, as it measures the ability of the detector to separate clean and backdoored adapters independently of a particular decision threshold. Accuracy is also reported for completeness. To isolate the quality of the learned representation from threshold calibration, the decision threshold is chosen using the known number of positive samples in the test set, i.e., the threshold is placed immediately after the \(n\)-th highest anomaly score, where \(n\) is the number of backdoored adapters. This allows us to evaluate the separability of the score distributions without conflating it with the threshold selection problem.

\section{Experiments}
\label{sec:experiments}
\paragraph{Datasets, Models, and Attacks.} We adopt PADBench~\cite{sun2025peftguard}, the backdoor adapter library introduced by PEFTGuard. PADBench is the largest library of backdoored PEFT adapters, comprising 13,300 LoRA adapters trained across multiple model families, fine-tuning tasks, and attack methods, with the query ($q$) and value ($v$) projection weights provided for each adapter. The base models include Llama-2-7B and Llama-2-13B~\cite{touvron2023llama}, Qwen1.5-7B~\cite{qwen1.5}, Flan-T5-XL~\cite{chung2024scaling}, and RoBERTa~\cite{liu2019roberta}. The fine-tuning tasks span two categories: task-specific datasets (IMDB~\cite{IMDB}, AG News~\cite{AG_news}, and SQuAD~\cite{rajpurkar2016squad}) and instruction-following datasets (toxic-backdoors-alpaca~\cite{toxic_backdoors_alpaca} and toxic-backdoors-hard~\cite{toxic_backdoors_hard}). Four backdoor attack methods are applied across the task-specific datasets: InsertSent~\cite{InsertSent}, Syntactic~\cite{Syntactic}, RIPPLES, and StyleBkd. In all cases, the detection task is binary: given only the adapter weights, determine whether the adapter is backdoored or clean.

\begin{table}
  \centering
  \setlength{\tabcolsep}{1mm}
  \begin{tabular}{@{}lcccc@{}}
    \toprule
    \makecell{Training/\\ \textit{Transfer}}
      & Sent
      & Ripple
      & Syn
      & Sty \\
    \midrule
    \textit{Sent}
      & \makecell{\textbf{.999}/.996\\ .405}
      & \makecell{\textbf{.829}/.670\\ .470}
      & \makecell{\textbf{.917}/.845\\ .457}
      & \makecell{\textbf{.882}/.718\\ .597} \\

    \textit{Ripple}
      & \makecell{\textbf{.903}/.851\\ .411}
      & \makecell{.998/\textbf{1.000}\\ .452}
      & \makecell{.776/\textbf{.828}\\ .464}
      & \makecell{\textbf{.894}/.795\\ .603} \\

    \textit{Syn}
      & \makecell{.880/\textbf{.892}\\ .365}
      & \makecell{.601/\textbf{.722}\\ .443}
      & \makecell{\textbf{1.000}/.998\\ .413}
      & \makecell{\textbf{.959}/.849\\ .624} \\

    \textit{Sty}
      & \makecell{.952/\textbf{.965}\\ .376}
      & \makecell{\textbf{.872}/.662\\ .447}
      & \makecell{\textbf{.996}/.986\\ .492}
      & \makecell{\textbf{1.000}/\textbf{1.000}\\ .640} \\
    \bottomrule
  \end{tabular}
  \caption{Zero-shot transfer on AG News attacks. In each cell, the first line reports \method{}/PEFTGuard AUROC and the second line reports WSD's AUROC. Bold marks the best result. Here, \textit{Sent}, \textit{Syn} and \textit{Sty} abbreviate for \textit{Insertsent}, \textit{Syntactic}, \textit{StyleBkd} respectively.}
  \label{tab:ag-news-zero-shot}
\end{table}

\paragraph{Experimental Infrastructure.}
All experiments were conducted on compute nodes equipped with 56 logical CPU cores and approximately 120~GiB of memory. Unless stated otherwise, all experiments used Llama-2 7B as the underlying model architecture.
The results on different architectures are available in the Supplementary Material.

We evaluate \method{} under detection settings of increasing difficulty. 
\paragraph{Zero-shot transfer between attacks on a known dataset.} The detector is trained on a single attack from \textit{InsertSent}, \textit{RIPPLE}, \textit{Syntactic}, and \textit{StyleBkd}, and is then evaluated on the remaining unseen attacks. This setting isolates whether the knowledge acquired from one attack transfers to other attacks on the same data distribution and compares with previously proposed methods. %

\paragraph{Leave-one-attack-out.} Here, the detector is trained jointly on multiple attacks from AG News and evaluated on an attack excluded from training. This experiment serves two purposes: (i) determining whether \method{} can exploit characteristics shared across different attacks, and (ii) evaluating whether its performance improves when the training set includes multiple attacks. Unlike the single-attack setting, this setup does not assume that a separate detector can be selected according to the attack encountered at test time. Instead, a single detector must generalize to an unseen attack after learning from several other attacks. %

\paragraph{Heterogeneous zero-shot.} In this setting, the evaluation pool spans AG News, IMDb, SQuAD, and Toxic Backdoors; six attack types (\textit{InsertSent}, \textit{RIPPLE}, \textit{Syntactic}, \textit{StyleBkd}, \textit{Alpaca}, and \textit{Hard}); five adapter methods (LoRA, AdaLoRA, DoRA, LoRA+, and QLoRA); and adapter ranks ranging from 8 to 2048. Each configuration represents a particular combination of dataset, attack, adapter method, and rank; the complete collection is listed in the first column of Table~\ref{tab:in-domain}.
The detector is trained across these heterogeneous configurations and evaluated on both (i) unseen attacks from a known dataset and (ii) known attacks instantiated using an unseen adapter rank or configuration. This setting measures whether the detector generalizes beyond individual attacks and training configurations, approximating a deployment scenario in which the provenance and configuration of a potentially malicious adapter are unknown. %

\paragraph{In-domain performance.} We also verify that our approach does not compromise on in-domain performance. Specifically, we evaluate the detector on attacks and configurations represented during training, including multiple datasets, attack types, adapter methods, and ranks. %

\subsection{Results}

\begin{table}
  \centering
  \setlength{\tabcolsep}{1mm}
  \begin{tabular}{@{}lcc@{}}
    \toprule
    \textbf{Held-out Attack}
      & \textbf{AUROC}
      & \textbf{ACC} \\
    \midrule
    \textit{InsertSent}
      & \textbf{.949}/.802/.366
      & \textbf{.931}/.776/.621 \\
    \textit{RIPPLE}
      & \textbf{.926}/.822/.445
      & \textbf{.893}/.813/.405 \\
    \textit{Syntactic}
      & \textbf{.910}/.850/.460
      & \textbf{.904}/.820/.661 \\
    \textit{StyleBkd}
      & \textbf{.989}/.884/.602
      & \textbf{.968}/.872/.667 \\
    \bottomrule
  \end{tabular}
  \caption{Leave-one-attack-out performance on AG News attacks. Each entry reports \method{}/PEFTGuard/WSD performance; bold marks the best result.}
  \label{tab:ag-news-loo-vs-24h}
\end{table}

\paragraph{Zero-shot detection on a known dataset.}
Table~\ref{tab:ag-news-zero-shot} reports the resulting known-attack-to-novel-attack transfer performance. \method{} outperforms PEFTGuard on 8 of the 12 off-diagonal transfer pairs, increasing the average AUROC from .8153 to .8718. This corresponds to an absolute improvement of 5.65 AUROC percentage points and a relative improvement of 6.93\%. By contrast, WSD performs near chance on average, achieving an off-diagonal AUROC of .4792 and being outperformed by \method{} on all 12 transfer pairs.

\paragraph{Leave-one-attack-out.} Table~\ref{tab:ag-news-loo-vs-24h} reports results for the multi-attack leave-one-out setting described above. 

Across all four held-out attacks, \method{} outperforms both PEFTGuard and WSD. It increases the mean AUROC from .8395 for the budgeted PEFTGuard model to .9433, corresponding to an absolute improvement of 10.38 AUROC percentage points. WSD performs near chance, with a mean AUROC of .4683, and trails PEFTGuard and \method{} on every held-out attack.

The budgeted PEFTGuard model does not surpass the best off-diagonal single-attack zero-shot result for any of the four target attacks. In contrast, the multi-attack \method{} model improves upon this reference for two attacks: by 5.35 percentage points on \textit{RIPPLE} and by 3.05 percentage points on \textit{StyleBkd}, giving an average improvement of 4.20 percentage points. On the remaining attacks, performance decreases by 0.33 percentage points on \textit{InsertSent} and 8.65 percentage points on \textit{Syntactic}, giving an average decrease of 4.49 percentage points. Across all four attacks, the mean AUROC therefore changes from .9448 to .9433, representing a small overall decrease of 0.15 percentage points relative to the best single-attack zero-shot results.

Overall, moving to the more realistic multi-attack training setting approximately preserves or improves \method{}'s performance on three of the four held-out attacks, while producing a substantial decrease on \textit{Syntactic}. These results suggest that the attacks share transferable characteristics that can be learned jointly, but also retain attack-specific properties that limit generalization in some cases.

\paragraph{Training and inference cost.} We compare the computational requirements of the three methods. 
PEFTGuard requires loading full checkpoints. For rank-256 Llama-2 7B LoRA adapters, the \texttt{float32} representation of each sample occupies approximately 4~GiB. Materializing the representations of 850 training samples requires approximately 3.3~TiB of RAM memory, which exceeded our hardware capacity. In comparison, the complete \method{} feature file for the same data occupies only 164.1~MiB, matching the theoretical reduction of $\sim$21{,}000$\times$. Since the PEFTGuard representations could not be loaded simultaneously, we load the checkpoints on-demand, which incurs in a significant dataloading overhead. %

\begin{figure}
    \centering
    \includegraphics[width=0.8\linewidth]{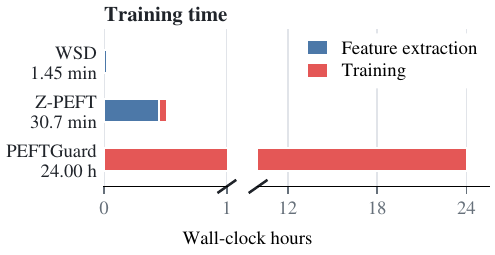}
    \includegraphics[width=0.8\linewidth]{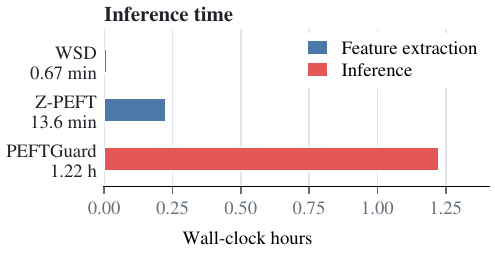}
    \caption{Timing decomposition for \method{}, WSD and PEFTGuard on the AG News leave-one-out experiment. Top: training time under a 24-hour PEFTGuard wall-clock budget. Bottom: inference time, separating preprocessing from the final prediction steps.}
    \label{fig:ag-news-timing}
\end{figure}
To limit the cost of the experiments, we fixed a maximum budget of 24 hours. PEFTGuard is the only method that requires the full budget, during which it completed almost three full epochs. As shown in Figure~\ref{fig:ag-news-timing} (Top), \method{}'s entire procedure finishes in 30.7 minutes, including 26.8 minutes for feature extraction and 3.9 minutes for feature selection and model fitting. WSD completes its corresponding procedure in only 1.45 minutes.

The inference-time comparison in Figure~\ref{fig:ag-news-timing} (Bottom) exhibits a similar pattern. \method{} requires 13.6 minutes for feature extraction and only 1.1 seconds for the subsequent classification step, whereas WSD requires 0.67 minutes and PEFTGuard approximately 1.22 hours. Considering the complete inference pipelines, \method{} is approximately 5.4 times faster than PEFTGuard. WSD is approximately 20.3 times faster than \method{} and 109.3 times faster than PEFTGuard.

\begin{figure}
    \centering
    \includegraphics{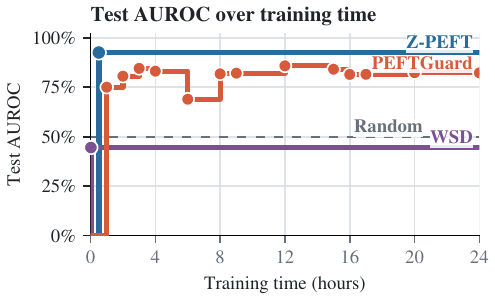}
    \caption{Test AUROC at fixed wall-clock checkpoints with \textit{RIPPLE} held out. \method{} and WSD complete their procedures in 30.7 and 1.45 minutes, respectively, whereas PEFTGuard is evaluated throughout a 24-hour training budget.}    
    \label{fig:ripple-auroc-time}
\end{figure}

Figure~\ref{fig:ripple-auroc-time} compares effectiveness at fixed wall-clock checkpoints when \textit{RIPPLE} is held out, while Table~\ref{tab:ag-news-loo-vs-24h} reports the complete attack-level comparison. WSD finishes substantially faster than \method{} (1.45 versus 30.7 minutes), but achieves an AUROC of only .445, below the random-ranking baseline of .5. In contrast, \method{} achieves an AUROC of .926, outperforming WSD by 48.08 AUROC percentage points while still completing in approximately half an hour.

\paragraph{Heterogenous zero-shot.}

Table~\ref{tab:llama2-7b-leave-one-out} reports leave-one-out performance across the heterogeneous configurations described above.
The AG News results are mixed when compared with the best off-diagonal single-attack zero-shot baseline. Performance improves by 6.85 AUROC percentage points on \textit{RIPPLE} and by 3.93 points on \textit{StyleBkd}, but decreases by 1.61 points on \textit{InsertSent} and by 11.85 points on \textit{Syntactic}. The detector nevertheless achieves a strong .887 AUROC on \textit{Hard-Rank256}, despite having observed this attack only at different ranks, indicating effective generalization across rank configurations. Conversely, performance on \textit{Alpaca} is relatively poor (.635 AUROC). Although \textit{Alpaca} shares its dataset with the \textit{Hard} attack, this result suggests that its distinct attack style limits transfer between the two attacks.

\begin{table}
  \centering
  \setlength{\tabcolsep}{1mm}
  \begin{tabular}{@{}lcc@{}}
    \toprule
    \textbf{Held-out Setting}
      & \textbf{AUROC}
      & \textbf{ACC} \\
    \midrule
    \textit{AG News InsertSent}
      & .936
      & .893 \\
    \textit{AG News RIPPLE}
      & .941
      & .915 \\
    \textit{AG News Syntactic}
      & .878
      & .861 \\
    \textit{AG News StyleBkd}
      & .998
      & .979 \\
    \textit{Alpaca}
      & .635
      & .640 \\
    \textit{Hard-Rank256}
      & .887
      & .832 \\
    \bottomrule
  \end{tabular}
  \caption{Heterogenous zero-shot performance for the Llama-2 7B detector. }
  \label{tab:llama2-7b-leave-one-out}
\end{table}

\paragraph{In-domain performance.} Table~\ref{tab:in-domain} reports performance on configurations seen during training, using a separate stratified test set containing only unseen samples. \method{} achieves a near-optimal performance, with an overall AUROC of .9986 and an accuracy of .9794. Thus, training on a heterogeneous collection preserves strong in-domain performance while supporting the zero-day generalization evaluated above.

\begin{table}
  \centering
  \small
  \setlength{\tabcolsep}{1mm}
  \begin{tabular}{@{}lcc@{}}
    \toprule
    \textbf{Dataset/Attack}
      & \textbf{AUROC}
      & \textbf{ACC} \\
    \midrule
    \textit{AG News--RIPPLE}
      & .998800 & .980000 \\
    \textit{AG News--InsertSent}
      & .983200 & .920000 \\
    \textit{AG News--StyleBkd}
      & 1.000000 & 1.000000 \\
    \textit{AG News--Syntactic}
      & .986800 & .940000 \\
    \midrule
    \textit{IMDb--RIPPLE}
      & .995600 & .960000 \\
    \textit{IMDb--InsertSent}
      & .992400 & .960000 \\
    \textit{IMDb--StyleBkd}
      & .980800 & .940000 \\
    \textit{IMDb--Syntactic}
      & .986400 & .940000 \\
    \midrule
    \textit{SQuAD--InsertSent}
      & 1.000000 & 1.000000 \\
    \textit{Toxic Backdoors--Alpaca}
      & 1.000000 & 1.000000 \\
    \midrule
    \textit{AdaLoRA--Hard ($r=8$)}
      & 1.000000 & 1.000000 \\
    \textit{DoRA--Hard ($r=256$)}
      & 1.000000 & 1.000000 \\
    \textit{LoRA+--Hard ($r=8$)}
      & 1.000000 & 1.000000 \\
    \textit{QLoRA--Hard ($r=256$)}
      & 1.000000 & 1.000000 \\
    \midrule
    \textit{Toxic Backdoors--Hard ($r=8$)}
      & 1.000000 & 1.000000 \\
    \textit{Toxic Backdoors--Hard ($r=16$)}
      & 1.000000 & 1.000000 \\
    \textit{Toxic Backdoors--Hard ($r=32$)}
      & 1.000000 & 1.000000 \\
    \textit{Toxic Backdoors--Hard ($r=64$)}
      & 1.000000 & 1.000000 \\
    \textit{Toxic Backdoors--Hard ($r=128$)}
      & 1.000000 & 1.000000 \\
    \textit{Toxic Backdoors--Hard ($r=256$)}
      & 1.000000 & 1.000000 \\
    \textit{Toxic Backdoors--Hard ($r=512$)}
      & 1.000000 & 1.000000 \\
    \textit{Toxic Backdoors--Hard ($r=1024$)}
      & 1.000000 & 1.000000 \\
    \textit{Toxic Backdoors--Hard ($r=2048$)}
      & .990000 & .900000 \\
    \midrule
    \textbf{Overall (pooled)}
      & .998572 & .979412 \\
    \bottomrule
  \end{tabular}
  \caption{In-domain detection performance.}
  \label{tab:in-domain}
\end{table}

\subsection{Feature Attribution Analysis}
We analyze the features used by \method{} through SHAP values~\cite{lundberg2017unified}. For our logistic regressor, each SHAP value represents a feature's additive contribution to the predicted log-odds, relative to the model's expected output. This contribution is determined by the corresponding model coefficient and by how much the observed feature value differs from its background mean.

We also analyze feature groupings based on transformer layer, attention head, projection matrix, and their combinations. Attribution across these architectural groupings is relatively uniform, with no individual layer, head, or projection matrix consistently dominating the predictions. Feature-family groupings instead reveal clearer and more attack-dependent attribution patterns. The top-5 feature-family attributions across the different sample groups are shown in Figure~\ref{fig:explainability}.
A further analysis on explainability is available in the Supplementary Material.

Overall, energy-concentration, kurtosis, entropy, and rank-based features account for most of the attribution magnitude. Their relative importance varies across attacks: for example, \textit{RIPPLE} relies primarily on energy-concentration and kurtosis, whereas stable rank is particularly important for the AdaLoRA \textit{Hard} attack. This suggests that the detector combines multiple complementary spectral properties rather than relying on a single universal signal.

\begin{figure}
    \centering
    \includegraphics{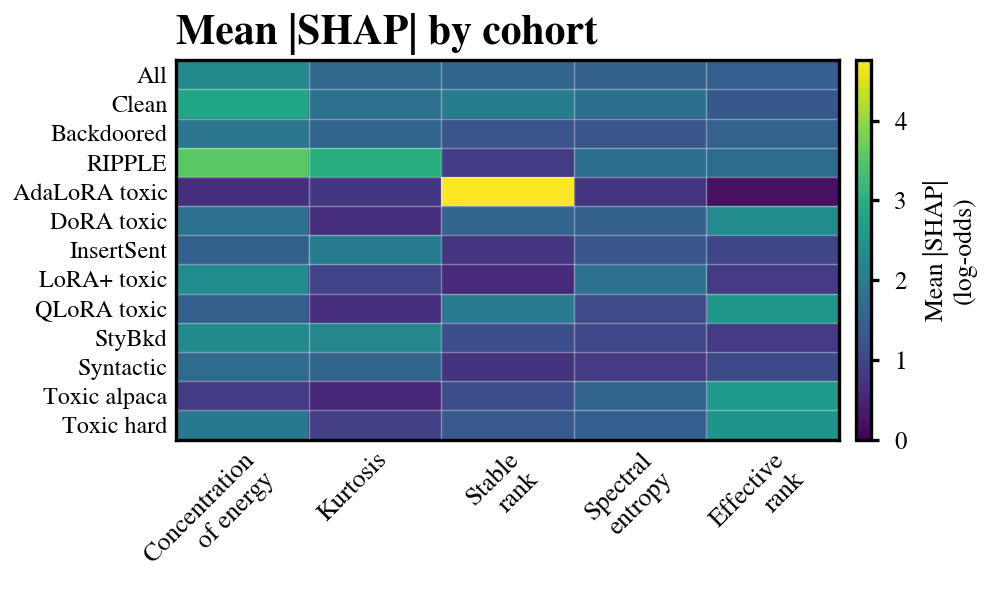}
    \caption{Grouped SHAP feature attribution across sample groups. Rows represent the test set, class-specific subsets, and individual attacks, while columns represent feature families. Each cell reports the mean absolute grouped SHAP value in log-odds space, computed by summing signed feature attributions within each family before averaging their magnitude. Higher values indicate a stronger contribution to the detector's predictions.}    
    \label{fig:explainability}
\end{figure}

\section{Limitations}
Although PADBench covers diverse attacks, datasets, architectures, and adapter configurations, it cannot capture the full variability of adapters encountered in practice. Consequently, performance may differ for naturally occurring backdoors or attack mechanisms that are substantially different from those represented in the benchmark. Moreover, PADBench provides adapter weights only for the query and value attention projections. Our evaluation therefore does not establish whether the observed spectral signatures generalize to adapters targeting other attention or MLP projections. While AUROC is threshold independent, the reported accuracy uses the known number of backdoored test samples to select a threshold and should therefore be interpreted as a measure of score separability rather than deployment-time accuracy; practical use would require threshold calibration under unknown attack prevalence. Finally, we do not evaluate an adaptive adversary that explicitly optimizes an adapter to mimic benign spectral statistics, making robustness to detector-aware attacks an important direction for future work.

\section{Conclusion}
We introduced \method{}, a scalable static detector that achieves state-of-the-art performance in the evaluated settings while being substantially faster and more memory-efficient than existing solutions. This scalability enables training on heterogeneous collections of datasets, attacks, PEFT methods, and adapter ranks, producing a single detector that nearly perfectly separates clean and backdoored adapters from known distributions while retaining strong generalization to zero-shot scenarios. \method{} bridges the main limitations of prior approaches: PEFTGuard preserves detailed structural information but operates on an input space too large for efficient multi-dataset training, whereas WSD is lightweight but relies on global projection-level spectral summaries that discard localized signatures. By combining compact spectral features with head-wise structural information, \method{} offers the best of both approaches and advances backdoor detection toward realistic open-world deployment.

\section*{Acknowledgments}
This paper has been partially supported by the CoEvolution project, funded by EU Horizon 2020 under GA n 101168559. We acknowledge ISCRA for awarding this project access to the LEONARDO supercomputer, owned by
the EuroHPC Joint Undertaking, hosted by CINECA (Italy).

\bibliography{aaai2027}

\appendix
\section{\method{} across architectures and PEFT configurations}

\paragraph{In-domain performance across architectures.}
We first evaluate \method{} across the model architectures available in PADBench. These experiments use the in-domain setting, where test adapters are drawn from configurations represented during training. The experimental support, however, is not identical across architectures. Flan-T5-XL, Qwen-1.5-7B, and Llama-2-13B share the same \textit{Toxic Backdoors--Hard}, rank-256, query-and-value-projection configuration and therefore provide the most direct comparison. RoBERTa-base is evaluated on IMDb adapters affected by InsertSent at rank~16, whereas Llama-2-7B is evaluated on a broader collection spanning multiple datasets, attacks, ranks, and adapter methods.

Table~\ref{tab:architecture-in-domain} shows consistently strong performance across all evaluated architectures. Flan-T5-XL and Qwen-1.5-7B achieve perfect separation, while Llama-2-13B obtains a comparable AUROC of .9980. Performance also remains high on the heterogeneous Llama-2-7B collection and on RoBERTa-base, with AUROC values of .9986 and .9908, respectively. These results indicate that \method{} is not tied to a single model family. Comparisons involving RoBERTa-base and Llama-2-7B should nevertheless be interpreted as architecture-specific in-domain results rather than as controlled comparisons of architecture alone, since their experimental support differs.

\begin{table}[!h]
  \centering
  \setlength{\tabcolsep}{1mm}
  \begin{tabular}{@{}lcc@{}}
    \toprule
    Architecture & AUROC & ACC \\
    \midrule
    Flan-T5-XL   & 1.0000 & 1.0000 \\
    Qwen-1.5-7B  & 1.0000 & 1.0000 \\
    Llama-2-13B  & .9980  & .9600 \\
    Llama-2-7B   & .9986  & .9794 \\
    RoBERTa-base & .9908  & .9400 \\
    \bottomrule
  \end{tabular}
  \caption{In-domain detection performance across model architectures.}
  \label{tab:architecture-in-domain}
\end{table}

\paragraph{Zero-shot generalization across ranks.}
We next evaluate whether \method{} generalizes to unseen adapter ranks through a leave-one-rank-out experiment on \textit{Toxic Backdoors--Hard}. For each run, adapters of one rank are excluded from detector training and used for evaluation, while the detector is trained on the remaining ranks.

\begin{table}
  \centering
  \setlength{\tabcolsep}{1mm}
  \begin{tabular}{@{}lcc@{}}
    \toprule
    Held-out Rank & AUROC & ACC \\
    \midrule
    $8$    & 1.0000 & 1.0000 \\
    $16$   & .9992  & .9800 \\
    $32$   & 1.0000 & 1.0000 \\
    $64$   & .9996  & .9800 \\
    $128$  & .9808  & .9800 \\
    $256$  & .9999  & .9960 \\
    $512$  & .9996  & .9798 \\
    $1024$ & 1.0000 & 1.0000 \\
    $2048$ & .9600  & .9200 \\
    \bottomrule
  \end{tabular}
  \caption{Leave-one-rank-out performance.}
  \label{tab:loo-rank}
\end{table}

Table~\ref{tab:loo-rank} shows near-optimal performance for held-out ranks from 8 to 1024, with AUROC never falling below .9808 and ACC remaining above .9798. The only notable degradation occurs at rank~2048, where \method{} nevertheless achieves an AUROC of .9600 and an ACC of .9200. Performance does not decrease monotonically with rank, indicating that the detector does not simply rely on rank-specific properties. Overall, the learned spectral signatures transfer effectively across ranks, although extrapolation to the largest evaluated rank is comparatively more challenging.

\paragraph{Zero-shot generalization across adapter methods.}
Finally, we assess whether \method{} generalizes to unseen PEFT methods through a leave-one-adapter-out evaluation. For each run, all adapters produced using one method are excluded from detector training and used for evaluation, while the remaining adapter methods constitute the training distribution.

As shown in Table~\ref{tab:loo-adapter}, \method{} transfers effectively to DoRA, LoRA+, LoRA, and QLoRA, achieving an AUROC of at least .9924 and an ACC of at least .9720. In contrast, performance drops sharply when AdaLoRA is held out, reaching an AUROC of .2628 and an ACC of .2880. The below-chance AUROC indicates that the learned scores largely reverse the ordering of clean and backdoored AdaLoRA adapters. This suggests that AdaLoRA's adaptive rank-allocation mechanism produces spectral characteristics that differ substantially from those of the other evaluated methods. Thus, \method{} generalizes strongly among closely related LoRA variants, but transfer is not guaranteed for methods with structurally different parameterizations.

\begin{table}[!h]
  \centering
  \setlength{\tabcolsep}{1mm}
  \begin{tabular}{@{}lcc@{}}
    \toprule
    Held-out Adapter & AUROC & ACC \\
    \midrule
    DoRA    & 1.0000 & 1.0000 \\
    LoRA+   & 1.0000 & 1.0000 \\
    LoRA    & .9956  & .9720 \\
    QLoRA   & .9924  & .9720 \\
    AdaLoRA & .2628  & .2880 \\
    \bottomrule
  \end{tabular}
  \caption{Leave-one-adapter-out performance.}
  \label{tab:loo-adapter}
\end{table}

\section{Extended Feature Attribution Analysis}

The main paper analyzes the spectral feature families used by \method{}. We extend that analysis along three complementary dimensions. First, we examine how attribution is distributed across layers, attention heads, projection blocks, and spectral feature families for different data cohorts. Second, we study the concentration of attribution over their joint cross-product. Finally, we inspect individual predictions to illustrate how these contributions combine into correct and incorrect decisions.

For a group of features \(G\), we define its attribution for a sample \(x\) as
\[
    \Phi_G(x) = \sum_{j \in G} \phi_j(x),
\]
where \(\phi_j(x)\) is the SHAP contribution of feature \(j\) in log-odds space. We measure the importance of \(G\) within a cohort \(\mathcal{C}\) by averaging the magnitude of its grouped attribution,
\[
    I_G(\mathcal{C})
    = \frac{1}{|\mathcal{C}|}
      \sum_{x \in \mathcal{C}} |\Phi_G(x)|.
\]
Summing signed contributions before taking their absolute value preserves cancellation between features belonging to the same group.

\begin{figure}
    \centering
    \includegraphics[width=\linewidth]{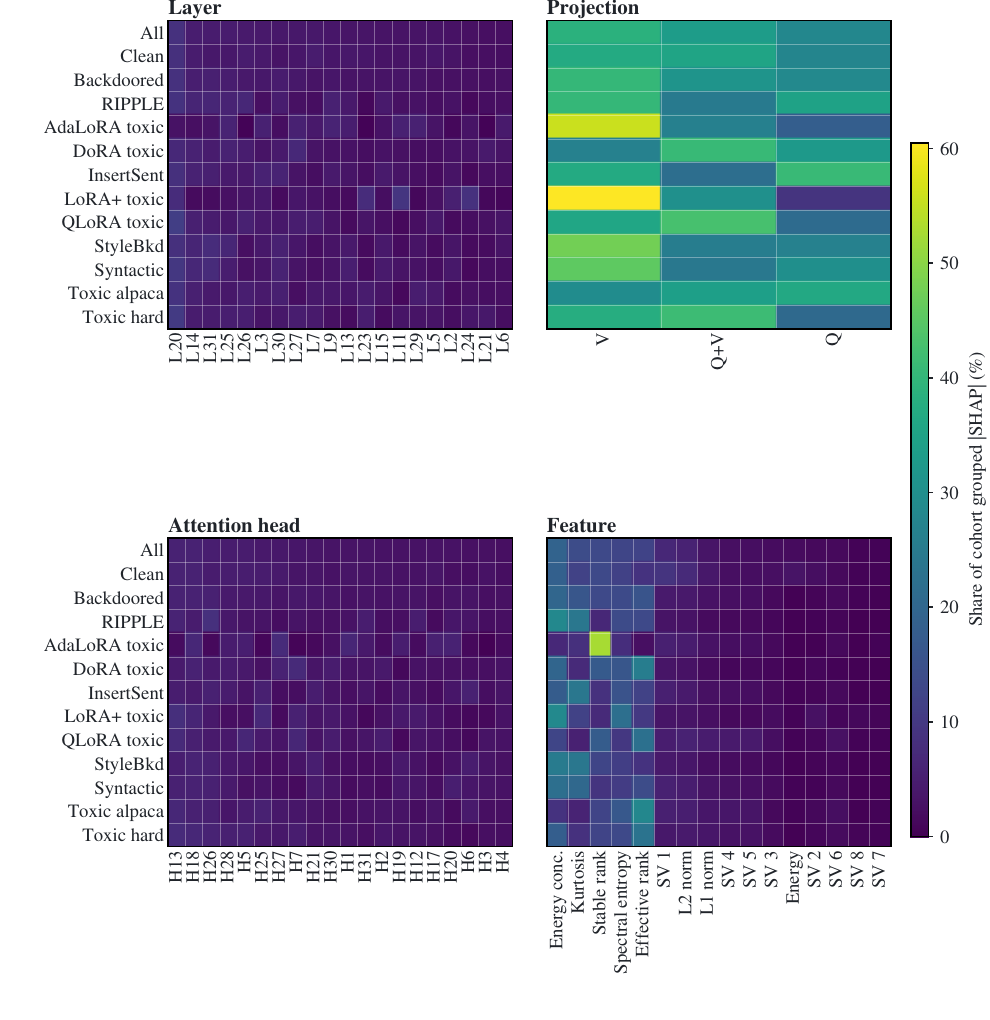}
    \caption{Cohort-level grouped SHAP attribution across transformer layers, projection blocks, attention heads, and spectral feature families. Signed SHAP contributions are summed within each group before taking the absolute value and averaging over the cohort. Each cohort row is normalized to \(100\%\); all panels use the same color scale.}
    \label{fig:shap-cohort-groupings}
\end{figure}

\paragraph{Cohort-level attribution structure.}
Figure~\ref{fig:shap-cohort-groupings} groups features separately by transformer layer, attention head, projection block, and spectral statistic. Within each grouping, cohort-level importance is normalized to sum to \(100\%\), allowing the relative distribution of attribution to be compared across clean, backdoored, and attack-specific subsets.

The results extend the feature-family analysis presented in the main paper. Energy concentration, kurtosis, spectral entropy, stable rank, and effective rank account for most of the attribution, whereas individual singular values and norm-based features generally contribute less. Their relative importance is nevertheless cohort dependent. RIPPLE relies strongly on energy concentration and kurtosis, while stable rank dominates the AdaLoRA cohort. This distinctive AdaLoRA profile is consistent with its adaptive rank-allocation mechanism and with the limited transfer observed when AdaLoRA is excluded from detector training.

Attribution is more diffuse across layers and attention heads. Although some locations receive greater importance for individual attacks, no single layer or head dominates consistently across cohorts. The projection-block analysis is more concentrated: the value projection accounts for the largest share globally and for several attack families, while the joint query--value and query projections become more prominent for particular cohorts. Thus, the detector relies on recurring spectral properties, but the architectural locations from which it obtains this evidence vary across attacks and adapter configurations.

\begin{figure}
    \centering
    \includegraphics[width=\linewidth]{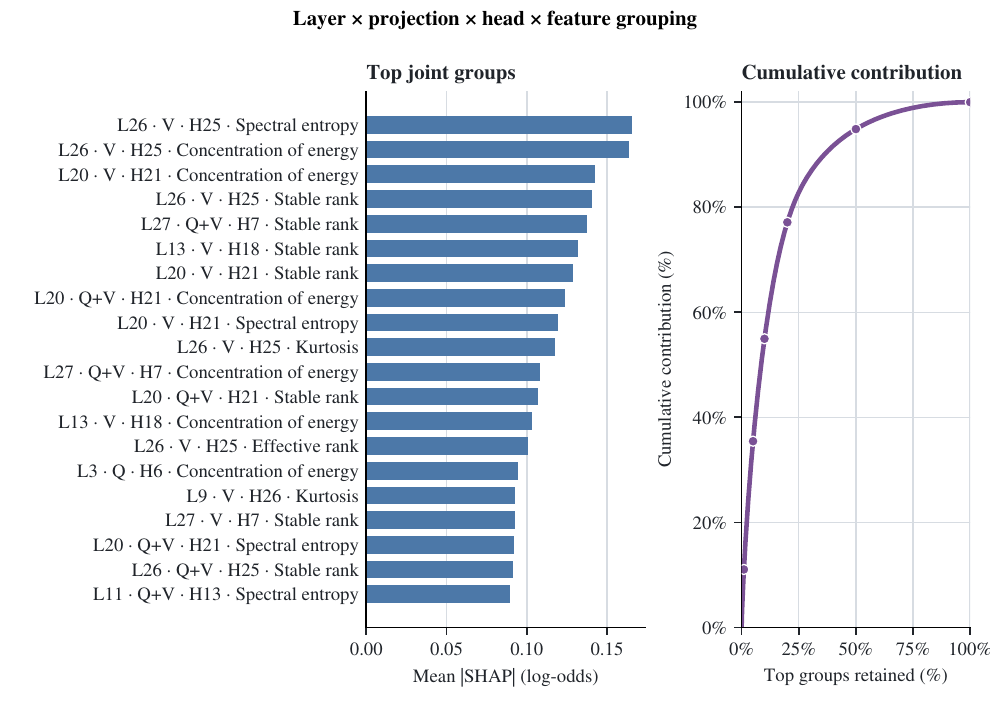}
    \caption{Global importance and concentration of joint layer--projection--head--feature groups. The left panel reports the groups with the largest mean absolute grouped SHAP attribution. The right panel shows the cumulative fraction of attribution captured by retaining an increasing percentage of groups.}
    \label{fig:shap-global-concentration}
\end{figure}

\paragraph{Concentration across joint feature groups.}
We next retain the complete grouping key defined by the cross-product of layer, projection block, attention head, and spectral statistic. Figure~\ref{fig:shap-global-concentration} reports the most important joint groups and the cumulative fraction of global attribution captured as progressively more groups are retained.

The highest-ranked groups predominantly combine value-projection features from layers 20 and 26 with spectral entropy, energy concentration, and rank-based statistics. Their individual mean absolute contributions remain small, however, with no single joint group exceeding approximately .17 log-odds. The cumulative analysis reveals moderate rather than extreme concentration: the top \(10\%\) of groups account for approximately \(55\%\) of total attribution, the top \(20\%\) account for approximately \(77\%\), and the top \(50\%\) account for approximately \(95\%\). Consequently, \method{} does not depend on one isolated architectural coordinate. Its decisions instead combine a relatively important subset of joint groups with a long tail of individually weak contributions.

\begin{figure}
    \centering
    \includegraphics[width=\linewidth]{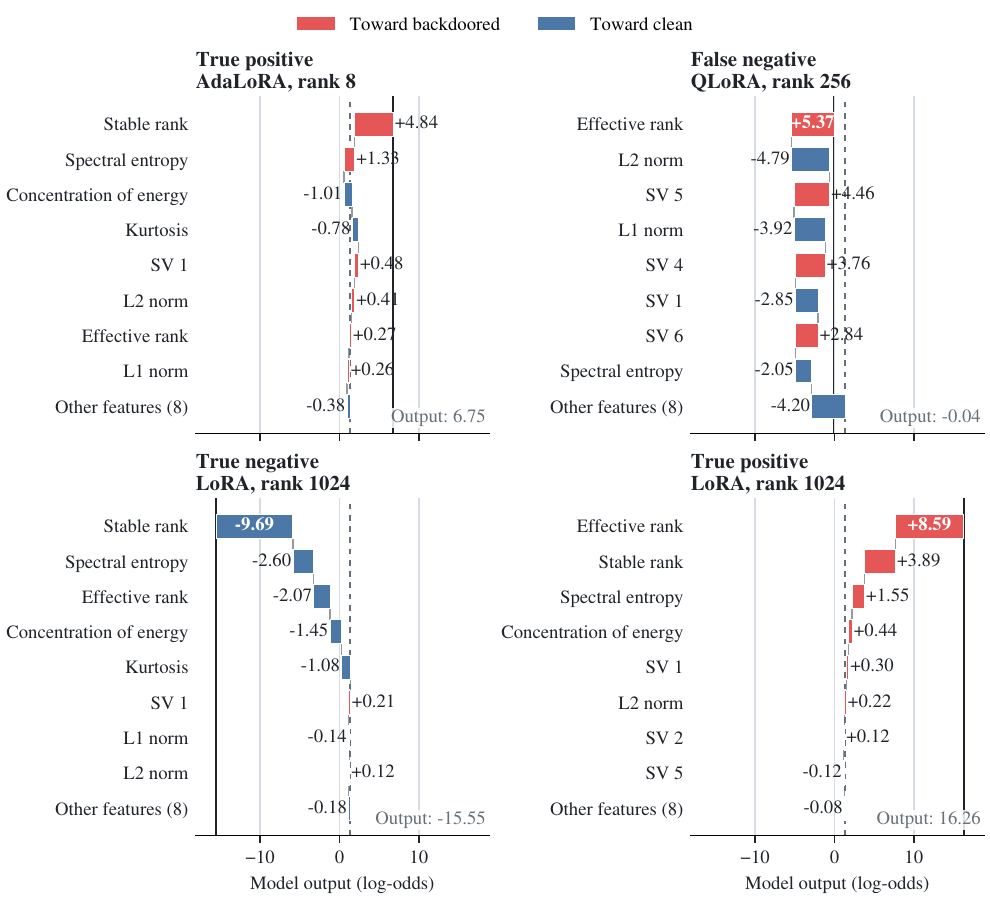}
    \caption{Local signed SHAP explanations for correctly and incorrectly classified adapters. Red bars move the prediction toward the backdoored class and blue bars toward the clean class. The dashed line denotes the expected model output \(E[f(X)]\), and the solid line denotes the final output \(f(x)\). ``Other features'' aggregates the remaining displayed feature-family contributions.}
    \label{fig:shap-local-explanations}
\end{figure}

\paragraph{Local decision explanations.}
Figure~\ref{fig:shap-local-explanations} presents signed SHAP contributions for four individual adapters. Positive contributions move the prediction toward the backdoored class, whereas negative contributions move it toward the clean class. Because the classifier output is expressed in log-odds, zero constitutes the decision boundary.

The correctly classified AdaLoRA adapter is driven primarily by stable rank, which contributes \(+4.84\) log-odds, followed by spectral entropy. This local explanation agrees with the cohort-level concentration of AdaLoRA attribution on stable rank. The paired LoRA rank-1024 examples further show that the same feature families can support opposite decisions. For the clean adapter, stable rank, spectral entropy, effective rank, energy concentration, and kurtosis all contribute toward the clean class, producing an output of \(-15.55\). For the backdoored adapter, effective rank, stable rank, and spectral entropy instead contribute strongly toward the backdoored class, producing an output of \(16.26\). The detector therefore does not interpret the presence of a particular feature family as intrinsically malicious; the direction and magnitude of its realized contribution determine the decision.

The QLoRA false negative illustrates a different behavior. Effective rank and several singular-value features provide substantial evidence for the backdoored class, but these contributions are offset by the \(L_1\) and \(L_2\) norms, spectral entropy, and other features. The resulting output is \(-.04\), immediately below the decision boundary. This example represents a borderline error caused by competing evidence rather than the complete absence of a backdoor signal.

Overall, the attribution analysis reveals a hierarchical decision process. A small set of spectral feature families recurs across cohorts, their evidence is distributed over multiple architectural locations, and individual predictions emerge from the signed combination of these contributions. These explanations characterize the behavior of the trained detector; they should not be interpreted as establishing a causal relationship between individual spectral properties and the presence of a backdoor.

\section{Classifier Hyperparameter Selection}

We select the \(L_2\)-regularized logistic-regression classifier through a grid search over the inverse regularization strength \(C\) and the class-weighting strategy. The search contains 12 candidates, obtained from all combinations of six values of \(C\) and two class-weight settings. Candidates are ranked by their mean cross-validation AUROC, and the configuration with the highest value is retained. Because \(C\) is the inverse regularization strength, smaller values correspond to stronger \(L_2\) regularization.

\begin{table}[t]
    \centering
    \setlength{\tabcolsep}{1mm}
    \begin{tabular}{@{}ll@{}}
        \toprule
        Hyperparameter & Values \\
        \midrule
        \(C\) &
        \(\{.001,.01,.1,1,10,100\}\) \\
        \texttt{class\_weight} &
        \(\{\texttt{None},\texttt{balanced}\}\) \\
        Solver & \texttt{lbfgs} \\
        Maximum iterations & \(5000\) \\
        Preprocessing & \texttt{StandardScaler} \\
        \bottomrule
    \end{tabular}
    \caption{Logistic-regression search space and fixed training settings.}
    \label{tab:logistic-regression-search}
\end{table}

\end{document}